\documentclass[10pt,twocolumn,letterpaper]{article}

\usepackage[compact]{titlesec}
\usepackage{iccv}
\usepackage{times}
\usepackage{epsfig}
\usepackage{graphicx}
\usepackage{amsmath}
\usepackage{amssymb}

\usepackage{booktabs}
\usepackage{adjustbox}
\usepackage{array}

\newcolumntype{R}[2]{%
  >{\adjustbox{angle=#1,lap=\width-(#2)}\bgroup}%
  l%
  <{\egroup}%
}

\graphicspath{{./Figures/}}
\usepackage{array}
\newcolumntype{M}{>{\centering\arraybackslash}m{65pt}}
\usepackage[font=small,labelfont=bf,up,textfont=up]{caption}
\usepackage{subcaption}
\usepackage{multirow}

\newcommand{\ig}[2]{\includegraphics[width=#1]{#2} } 
\newlength{\sm}
\usepackage[pagebackref=true,breaklinks=true,letterpaper=true,colorlinks,bookmarks=false]{hyperref}

\iccvfinalcopy 

\def\iccvPaperID{1781} 

\ificcvfinal\fi
\begin{document}

\title{Quality Resilient Deep Neural Networks}

\author{Samuel Dodge and Lina Karam\\
Arizona State University\\
{\tt\small \{sfdodge,karam\}@asu.edu}
}

\maketitle

\begin{abstract}
  We study deep neural networks for classification of images with quality distortions. We first show that networks fine-tuned on distorted data greatly outperform the original networks when tested on distorted data. However, fine-tuned networks perform poorly on quality distortions that they have not been trained for. We propose a mixture of experts ensemble method that is robust to different types of distortions. The ``experts'' in our model are trained on a particular type of distortion. The output of the model is a weighted sum of the expert models, where the weights are determined by a separate gating network. The gating network is trained to predict optimal weights for a particular distortion type and level. During testing, the network is blind to the distortion level and type, yet can still assign appropriate weights to the expert models. We additionally investigate weight sharing methods for the mixture model and show that improved performance can be achieved with a large reduction in the number of unique network parameters.
\end{abstract}

\section{Introduction}
Neural networks can be thought of as a sequence of layered operations that attempt to model an arbitrary function $f(x) = y$ given many examples of $x$ and $y$. A network's parameters are optimized given a large amount of training data $x \in X$ and corresponding target values $y \in Y$. However in many practical situations, the testing data may not be drawn from the same distribution as $X$, but rather from $\widetilde{X}$ which is a distorted version of $X$. In computer vision problems, the distortion could be an artifact of the camera (e.g., noise or blur), or could be a result of the environment (e.g., rain or fog). It is even possible to intentionally generate a $\widetilde{X}$ that looks similar to the original data but is consistently misclassified \cite{adversarial2}.

If it is known a-priori that the testing data may lie outside of the distribution of the training data, a simple solution is to re-train (or fine-tune) the network on data from the known distorted distribution. This has been shown to achieve good performance in the case of noisy \cite{distort-icassp} or blurry data \cite{blurnetworks}. The network is able to adapt to the new distribution and give more correct predictions.

\begin{figure}[!tb]
  \centering
  \includegraphics[width=0.45\textwidth]{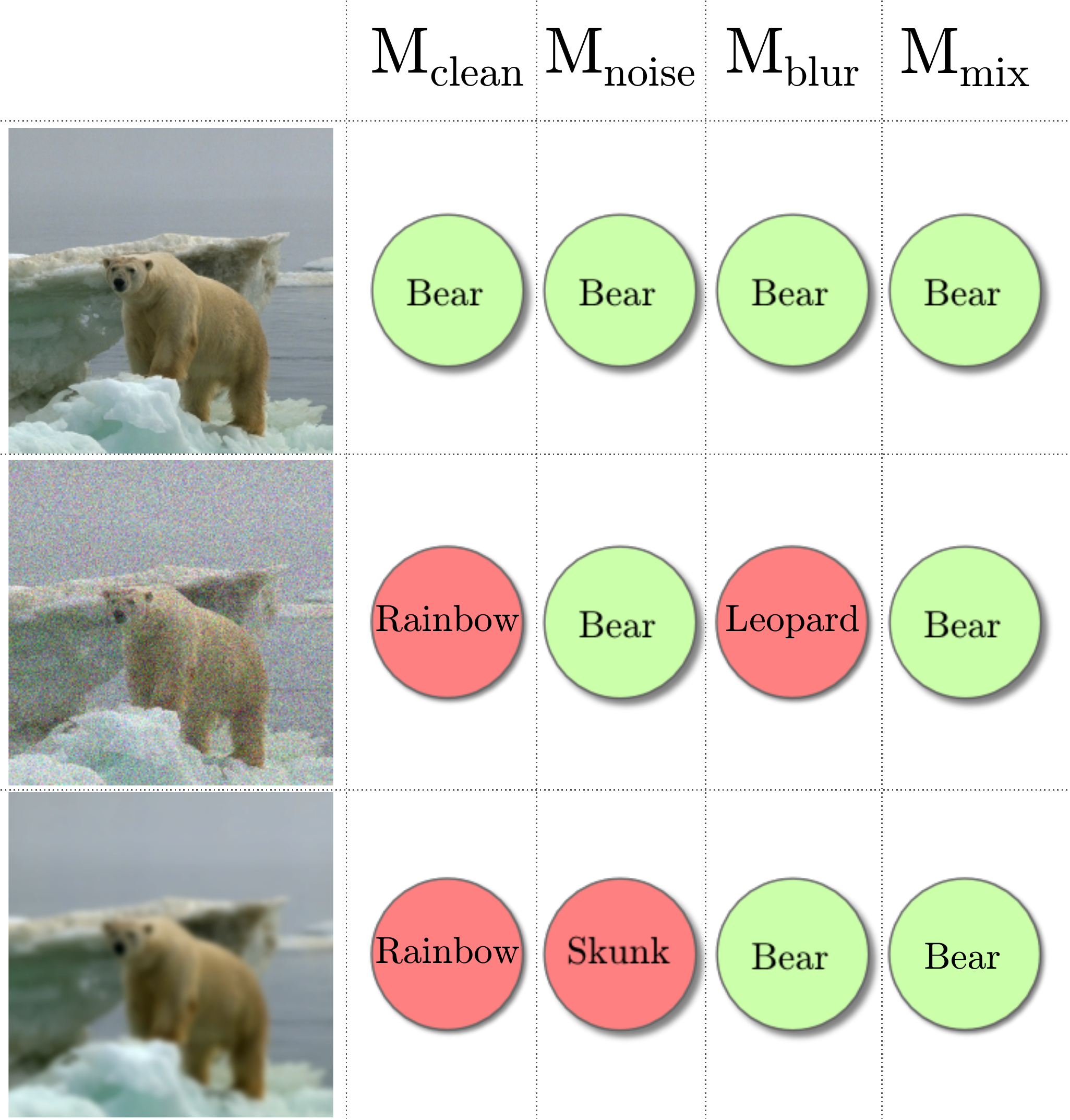}
\caption{\textbf{Example performance of quality resilient networks on various quality distortions.} This table shows the class prediction for an image under several different types of distortions (from top to bottom: clean, Gaussian noise and Gaussian blur). The original VGG16 network ($M_{clean}$) fails on distorted images. Networks fine-tuned on different types of distortions perform well on that particular distortion, but not on other distortion types ($M_{noise}$ and $M_{blur}$). Our mixture of experts based model ($M_{mix}$) performs well over all distortion types as well as the original clean image.}
\label{fig:intro}
\end{figure}

In this paper, we consider the scenario where there are multiple distortions. A network trained on one distortion $\widetilde{X}_1$ may not generalize when tested on distortion $\widetilde{X}_2$, and vice-versa. One solution is to train a model on the union distribution $\widetilde{X}_1 \cup \widetilde{X}_2$, but we show that this does not perform better than models trained on a single distortion type.

Given that networks can adapt to a particular distortion type, but not multiple types at once, we propose a mixture of experts model \cite{mixture} where each expert is trained on a particular distortion type. The mixture can more accurately model $\widetilde{X}_1$ and $\widetilde{X}_2$ simultaneously than a single model trained on $\widetilde{X}_1 \cup \widetilde{X}_2$. We first show that a simple averaging ensemble performs better than models trained on single distortions. Next we show that adding weights to the ensemble gives the best performance across several different types of image quality distortions. A separate gating network determines the weight to assign to each ensemble member. Finally, we note that the limitation of ensemble methods is the increased number of model parameters, and propose parameter sharing techniques that achieve improved performance with less parameters than the full ensemble method.

We consider two common image distortions: additive Gaussian noise and Gaussian blur, although our method could be applicable to other distortion types. Figure \ref{fig:intro} shows an example image corrupted with these distortions and the class prediction of several models. The models trained on particular types of noise ($M_{noise}$ and $M_{blur}$) perform well on the trained distortion type, but fail to generalize to other distortions. By contrast, our mixture model ($M_{mix}$) predicts the correct class under all distortion types.

\paragraph{Related Works}
Dodge and Karam \cite{dodge} showed that deep neural networks trained on clean data perform poorly when tested on low quality distorted images. This performance decrease is seen for several different network architectures. Noise and blur distortions cause the greatest degradation in network performance. Comparatively, JPEG and contrast distortions do not significantly affect networks. The VGG16 architecture \cite{vgg} was shown to be more robust compared with AlexNet or GoogleNet architectures.

A simple approach to add robustness to neural networks is to fine-tune the network on images with the expected distortions. Vasiljevic \etal \cite{blurnetworks} show that this approach works well for blurred images. They also show that to achieve good performance on clean and blurred images, it is best to train on data consisting of half clean and half blurred images. Similarly Zhou \etal \cite{distort-icassp} show the effectiveness of fine-tuning for both noisy images and blurred images. Interestingly, the model trained on both noise and blur has a much higher error rate than the average error rate of models trained only on noise and blur when these latter models are tested on their respective distortions.

Diamond \etal \cite{dirty-pixel} describe a system that prepends a neural network with additional layers that serve to undistort the image. For these undistorted images, additional fine-tuning of a deep neural network is performed. The method assumes knowledge of camera noise and blurring parameters, but for general applications (e.g., images from the Internet) the camera parameters may not be available. This limitation greatly limits the applicability of the method.

BANG training \cite{bang} is a method for training neural networks that are less susceptible to noisy input data. In BANG training, gradients in a mini-batch are scaled to give larger gradients to already correctly classified samples. Consequently, the correct samples are pushed further from the decision boundaries. Thus the resulting network is less susceptible to noise, which would otherwise push samples over decision boundaries. Fine-tuning with noisy samples may also push the samples away from decision boundaries, so it is not clear if BANG training offers any advantage compared with a simple fine-tuning baseline.

Stability training \cite{stability} is another training methodology designed to learn a more robust network. The method trains a network with noisy images and tries to match the soft-max outputs from an identical network trained with clean images. Stability trained networks were shown to have greater invariance to JPEG compression than traditionally trained networks. However, the usefulness of this is limited because existing networks are already quite robust to JPEG distortions as shown in \cite{dodge}.

Distortions can also be generated in an adversarial manner such that they are imperceptible to humans, but cause neural networks to perform poorly \cite{adversarial2, deepfool}. There are several approaches that attempt to alleviate susceptibility to adversarial samples. Networks can simply be re-trained on these adversarial samples to obtain more robustness \cite{adversarial1}. Another technique uses soft-max distillation training (similar to \cite{stability}) to achieve robustness. Although adversarial samples represent a potential security problem for neural networks, they are not as prevalent as images distorted with common image quality distortions.

Many of the aforementioned approaches offer solutions to problems where there is a single type of distortion. In this work we propose a method that can perform well on different types and levels of distortions. We test on blur and noise because they have opposite characteristics. Noise injects high frequency information into the image, and blur removes existing high frequency information. We first perform baseline experiments using fine-tuning on the distortions similar to the experiments in \cite{blurnetworks,distort-icassp}. Next, we propose our mixture of experts model and show that it achieves better performance than the fine-tuned models and other ensemble baselines. Finally, we perform weight sharing experiments to show that the mixture model can achieve better performance without significantly increasing the number of parameters over a single model.

\section{Baselines using Fine-tuning}
\label{sec:finetune}
In fine-tuning, the network is allowed to adapt to the statistics that are present in distorted images. In this section we establish fine-tuning baselines on blurred and noisy images.

\begin{figure*}[!tb]
  \centering
  \includegraphics[width=0.98\textwidth]{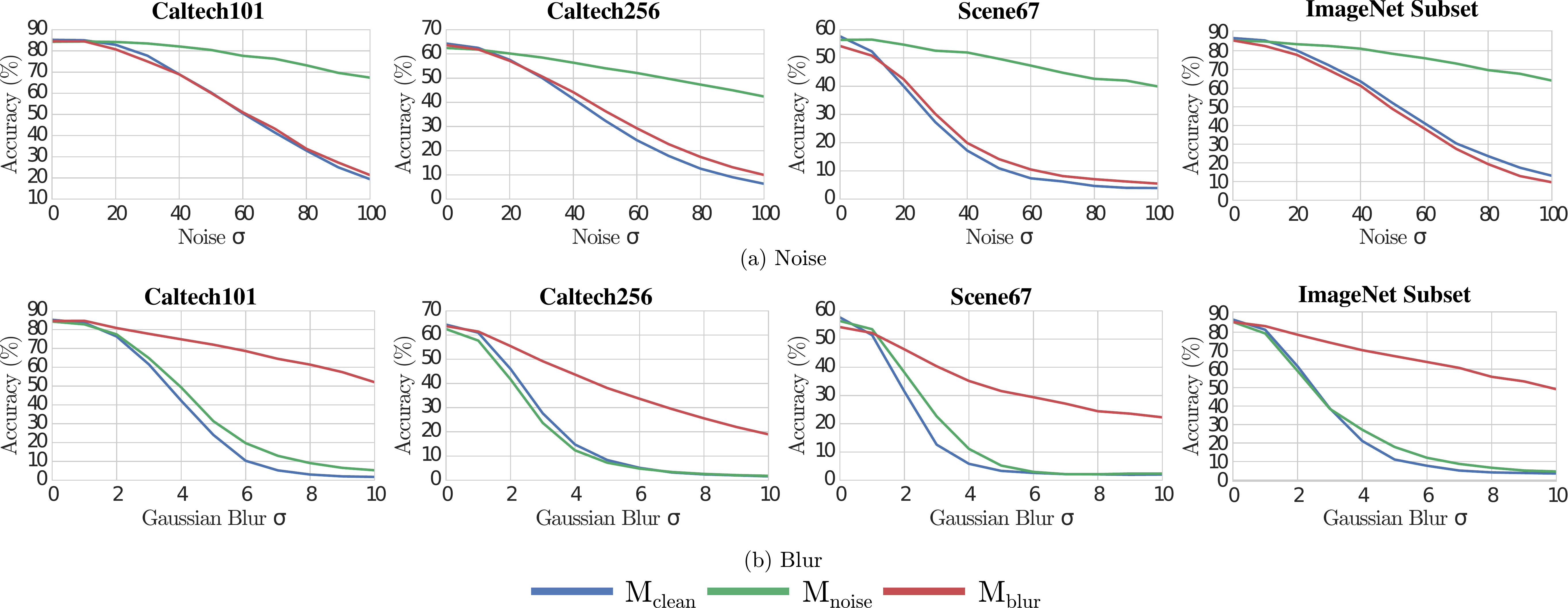}
  \caption{\textbf{Fine-tuning results.} Models trained with different types of distortions are tested on noisy images (a), and blurred images (b). $M_{noise}$ performs well on noisy images, but performs poorly on blurred images. Similarly, $M_{blur}$ performs well on blurry images but poorly on noisy images.}
  \label{fig:accuracy}
\end{figure*}

\paragraph{Distortions}
We test two types of distortions: additive Gaussian noise and Gaussian blur. For noise we add random Gaussian noise to each pixel with a standard deviation $\sigma_n$ ranging from 0 to 100. For blur we apply Gaussian blur with a standard deviation $\sigma_b$ ranging from 0 to 10. 

\paragraph{Datasets}

We focus our experiments on four medium sized datasets that consist of high resolution images: Caltech101 \cite{caltech101}, Caltech256 \cite{caltech256}, Scene67 \cite{scene67}, and a 50-class subset of the ImageNet (ILSVRC2012) dataset \cite{imagenet}. The Caltech datasets contain images of common objects, and the Scene67 dataset consists of images of indoor scenes. For the ImageNet dataset, we randomly choose 50 classes from the 1000 available classes. This makes computation time more feasible.

For both the Caltech101 and Caltech256 datasets we follow the standard protocol and randomly take 25 images per class for training, 5 per class for validation, and up to 50 images per class for testing. For the Scene67 dataset, we used the given training and testing split. For validation, 20\% of the given training images are held out during training. For the ImageNet dataset we use the given validation data for testing and split the given training data into 80\% for training and 20\% for validation.

\paragraph{Deep Neural Network}
In our experiments, we use the VGG16 model \cite{vgg}, as it was found to be the most resilient to distortions in \cite{dodge}. For each dataset, we replace the last fully connected layer with a new layer with the number of units corresponding to the number of classes in the dataset. The new layer is initialized using Xavier initialization \cite{glorot}. All other layers are initialized with the values trained on the full ImageNet dataset \cite{imagenet}.

\paragraph{Fine-tuning strategy}
We train the models using stochastic gradient descent with momentum using a learning rate of 0.001 for all the layers except for the last, and a learning rate of 0.01 for the last layer. The last layer is the only layer that has not been pre-trained, so it requires a larger learning rate. The learning rates are small because we do not want the fine-tuning to deviate too far from the pre-trained weights and over-fit the new datasets. We use a mini-batch size of 32 images. We do not perform any weight decay or learning rate scheduling. We train for as many epochs until the validation loss stops decreasing.

First we train a model using the non-distorted training images from the dataset. We denote this model $M_{clean}$. Next we train two additional models: $M_{noise}$ is fine-tuned on noisy images and $M_{blur}$ is fine-tuned on blurry images. The $M_{noise}$ and $M_{blur}$ models are initialized with the weights from the $M_{clean}$ model. During training, half of the images in a mini-batch are distorted with a random distortion level. For blur, this random level is the standard deviation of the Gaussian blur kernel and is randomly sampled from a uniform distribution $\sigma_b \sim U(0,10)$. The noise parameter is the standard deviation of the additive Gaussian noise and is randomly sampled from a uniform distribution $\sigma_n \sim U(0,100)$. Applying distortion to only half of the mini-batch images allows the model to perform well on both clean and distorted images (as shown in \cite{blurnetworks}). The distortion is applied on-line during training so that each sample in each mini-batch has unique random distortion levels.

We resize each image to $256 \times 256$ pixels and then take the $224 \times 224$ pixel central crop (as in \cite{vgg}). After resizing and cropping we apply a distortion if desired. Then we subtract the mean of the full ImageNet dataset used to train the original VGG16 model.


\paragraph{Results}
For each model and distortion type, we test 10 levels of distortions for each testing image (Figure~\ref{fig:accuracy}). An important consideration is how well a model performs on both clean images and distorted images. An ideal curve would be flat, with equal performance on distorted images and clean images.

Another consideration is \emph{cross-distortion} performance. An ideal model would achieve good performance on all types of distortions. However, we see that the model trained on noisy images performs poorly on blurred images. Likewise, the model trained on blurred images performs poorly on noisy images.

\begin{figure}
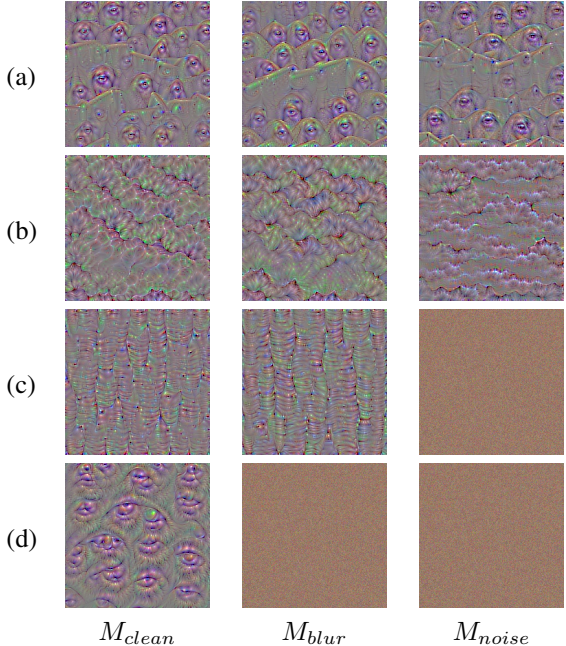

\setlength{\tabcolsep}{1pt}
\centering
     \begin{tabular}{m{15pt}MMM}
     (a) &
     \ig{\sm}{filters/clean_conv5_1_filter_30.png} &
     \ig{\sm}{filters/blur_conv5_1_filter_30.png} &
     \ig{\sm}{filters/noise_conv5_1_filter_30.png} \\
     (b) &
     \ig{\sm}{filters/clean_conv5_1_filter_10.png} &
     \ig{\sm}{filters/blur_conv5_1_filter_10.png} &
     \ig{\sm}{filters/noise_conv5_1_filter_10.png} \\
     (c) &
     \ig{\sm}{filters/clean_conv5_1_filter_14.png} &
     \ig{\sm}{filters/blur_conv5_1_filter_14.png} &
     \ig{\sm}{filters/noise_conv5_1_filter_14.png} \\
     (d) &
     \ig{\sm}{filters/clean_conv5_1_filter_12.png} &
     \ig{\sm}{filters/blur_conv5_1_filter_12.png} &
     \ig{\sm}{filters/noise_conv5_1_filter_12.png} \\
     & $M_{clean}$ & $M_{blur}$ & $M_{noise}$ \\
     \end{tabular}
  \caption{\textbf{Visualizations of neurons in the Conv5\_1 layer.} We compute the input that maximizes the response of selected neurons from the Conv5\_1 layer in three fine-tuned networks. Some neuron characteristics are relatively unchanged (a). Other neurons show some adaptation (b). Other neurons are effectively ``turned off'' by fine-tuning (c and d), so the visualization optimization cannot find a good input to maximize the response.}
\label{fig:filter}
\end{figure}

For these two types of distortions it seems that the fine-tuned models can specialize to one particular type of distortion, but cannot generalize to the other type of distortion. We can visually see this effect by analyzing the filter responses of the learned networks. We use the idea of Erhan \etal\cite{visualize} and compute inputs that maximally activate a single neuron. We start with an image with random pixel values and use gradient ascent to change the pixel values of the image to increase the activation of a chosen neuron. The resulting image indicates what kind of patterns the neuron responds to. Figure \ref{fig:filter} shows these visualizations for several neurons from the Conv5\_1 for the $M_{clean}$, $M_{blur}$, and $M_{noise}$ models. For some neurons the response is similar to the clean model, but the responses of other neurons are noticeably different. Furthermore, some neurons are completely turned off such that the gradient ascent optimization cannot find inputs that increase the activation from the starting random initialization. This indicates that during fine-tuning, the networks may turn off some neurons to achieve robustness.

\section{Ensembles of Networks}
\label{sec:ensemble}
Given that the individually trained networks do not achieve good performance across all levels and types of distortions, it may be useful to consider using ensembles of the individual models. We first define a matrix that consists of the outputs of the models trained on distortions from Section \ref{sec:finetune}:
\begin{equation}
  \mathbf{P}(x) = 
\left( \begin{array}{c}
    \mathbf{p}(x | M_{clean}) \\
    \mathbf{p}(x | M_{noise}) \\
    \mathbf{p}(x | M_{blur}) \\
  \end{array} \right)
  \end{equation}
\noindent where $\mathbf{p}(x | M)$ is a row vector of soft-max outputs of the last layer of the model $M$ for the input image $x$. 


\paragraph{Averaging Ensemble}
As a baseline we form an averaging ensemble from the individually trained networks. We use the networks trained in the previous section on clean, noisy, and blurred images, and average the outputs of the soft-max functions as:
\begin{equation}
  \mathbf{p}(x | M_{avg}) = \frac{1}{n} \mathbf{1}^T \mathbf{P}(x)
\end{equation}

\noindent where $n$ is the number of models and $\mathbf{1}$ is an $n$-length column vector of all 1s.

This averaging model is conceptually similar to the multi-column neural network \cite{multicolumn}. However the multi-column network assumes clean inputs and artificially distorts the data before passing it to each column network. Here we assume that the distortion type and level are unknown and pass the same input to all of the ensemble members.

\paragraph{Mixture of Experts}
A second approach is to assign weights to the output of each ensemble member. This approach can be modeled as a mixture of experts paradigm \cite{mixture}. In this case the three experts are the $M_{clean}$, $M_{noise}$, and $M_{blur}$ models. For a given input image $x$ we compute a vector $\mathbf{w}$ that determines the mixture weights. The output probabilities of the mixture model $M_{mix}$ are computed as:
\begin{equation}
  \mathbf{p}(x | M_{mix}) = \mathbf{w}(x)^T \mathbf{P}(x)
\end{equation}

\begin{figure*}[!tb]
  \centering
  \includegraphics[width=0.95\textwidth]{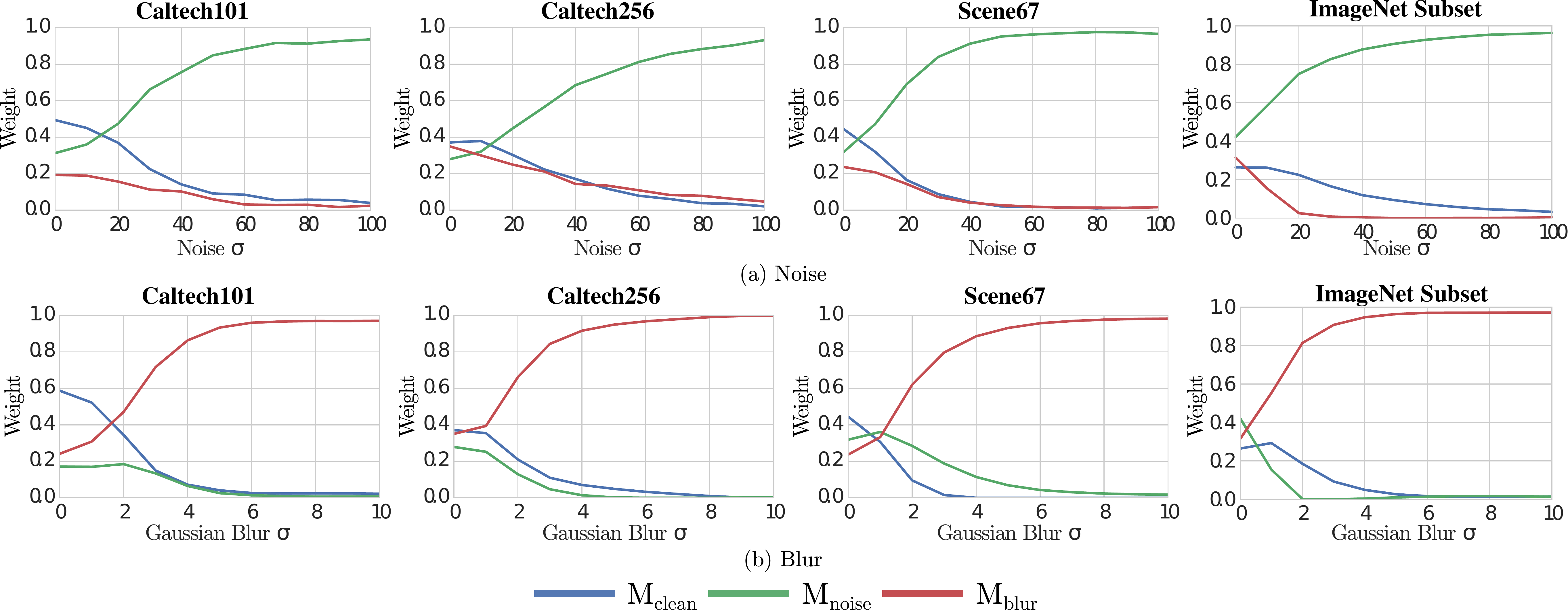}
  \caption{\textbf{Optimal weights.} We compute the optimal weighting function of the three models to achieve the minimum categorical cross-entropy. These optimal weights are used as the target to train a gating network that can predict weights for an image with unknown distortion type and level.}
  \label{fig:weights}
\end{figure*}

Our goal is to build a gating network that can determine the optimal weights for any given input image. The gating network does not necessarily give all of the weight to a single model, because there can be an ensembling benefit by combining multiple models. In the case where all elements of $\mathbf{w}$ are equal, the mixture model is equivalent to the averaging ensemble.

The ``ground-truth'' optimal weights must first be determined using the validation data so that the gating network can later be trained. We use the validation data instead of the training data because the training data is already well fit by the ensemble members. It is possible to compute the optimal weights on a per-image basis, however we found that predicting these weights is difficult, because the weights may vary between images with the same distortion type and level. Instead we consider all images with a particular distortion type and distortion level, and compute the optimal weights for this set of images. The ground-truth weights are computed by minimizing the categorical cross-entropy:
\begin{equation}
  \label{eq:minweight}
  \begin{aligned}
    & \underset{\mathbf{w}_{d,v}}{\text{min}}
    & &  - \sum_{x \in \mathbb{I}_{d,v}}\sum_j{t_{x,j} \log(\textbf{w}_{d,v}^T \textbf{P}(x)_j)} \\
    & \text{s.t.}
    & & ||\mathbf{w}_{d,v}||_1 = 1
  \end{aligned}
  \end{equation}

\noindent where $\mathbb{I}_{d,v}$ is a set of images with distortion type $d$ and distortion level $v$. $t$ is a binary value that is equal to one if image $x$ is of class $j$. $\mathbf{P}(x)_j$ corresponds to the $j$th column of $\mathbf{P}(x)$. Finally $\mathbf{w}_{d,v}$ is the optimal weight vector we wish to find. The constraint ensures that the weights sum to one.

We minimize Eq. \ref{eq:minweight} using the SLSQP algorithm \cite{slsqp} and the validation data of each dataset. Figure \ref{fig:weights} shows the optimal weights for the different distortion levels. For low distortion levels, relatively uniform weights give the best performance. This roughly corresponds to an averaging ensemble. For larger distortion levels, the optimal mixture favors the model learned on that particular distortion type. 


\begin{table}[h]
  \centering
  \footnotesize
  \begin{tabular}{ll}
    \toprule
    Layer & Hyper parameters\\
    \midrule
    Input & 224 $\times$ 224 pixels \\
    Conv-1 & 16 (3 $\times$ 3) filters \\
    Pool-1 & 2 $\times$ 2 max pooling \\
    Conv-2 & 32 (3 $\times$ 3) filters \\
    Pool2 & 2 $\times$ 2 max pooling \\
    Conv3 & 64 (3 $\times$ 3) filters \\
    Pool3 & 2 $\times$ 2 max pooling \\
    Conv-4 & 64 (3 $\times$ 3) filters \\
    Pool4 & 2 $\times$ 2 max pooling \\
    Conv-5 & 64 (3 $\times$ 3) filters \\
    Conv-6 & 16 (1 $\times$ 1) filters \\
    Conv-7 & 3 (1 $\times$ 1) filters \\
    Global Average Pooling & \\
    \bottomrule
  \end{tabular}
  \caption{Parameters of the gating network.}
  \label{tab:gating}
  \end{table}

Next we train a gating network to predict these optimal weights. Instead of using a VGG16 network, we train a much smaller network. The gating network consists of 7 convolutional layers and a global average pooling output layer (Table \ref{tab:gating}). The last element of the network is a soft-max non-linearity, which ensures that the outputs sum to one and the values of the outputs are always between zero and one. The total number of network parameters is 98.5 thousand, which is negligible compared to the 135 million parameters of each of the VGG16 ensemble members. The network is trained on the training images with random distortion types and levels. The target weight is taken from the weights learned in Eq. \ref{eq:minweight} for the particular distortion level and type. The gating network minimizes a regression loss to predict the weights for a given image:
\begin{equation}
  \begin{aligned}
    & \underset{\mathbf{g}}{\text{min}}
    & &  \sum_{d \in \mathbb{D}} \sum_{v \in \mathbb{V}_d} \sum_{x \in \mathbb{I}_{d,v}} ||\mathbf{g}(x_{d,v}) - \mathbf{w}_{d,v}||_2^2 + \lambda ||\mathbf{g}(x_{d,v})||_2^2 \\
  \end{aligned}
\end{equation}

\noindent where $\lambda$ is a regularization parameter, $\mathbb{D}$ is the set of all distortion types, $\mathbb{V}_d$ is a set of possible distortion levels for distortion $d$, and $\mathbf{g}(x)$ is the output of the gating network. In our experiments we set $\lambda = 0.01$. We can obtain similar results with different $\lambda$ values, and reach similar conclusions from our experiments.

The gating network is trained with stochastic gradient descent using the same parameters as the training procedure for the ensemble members. Half of the images in each mini-batch are undistorted, one quarter of the images are blurred with random levels, and one quarter of the images have noise with random levels. The levels are chosen randomly as in Section \ref{sec:finetune}.

Note that each element of the mixture model (the ensemble members and the gating network) are trained independently. Some benefit could be gained by joint training the ensemble, but due to memory constraints we train each element independently.

\begin{figure*}[!tb]
  \centering
  \includegraphics[width=0.97\textwidth]{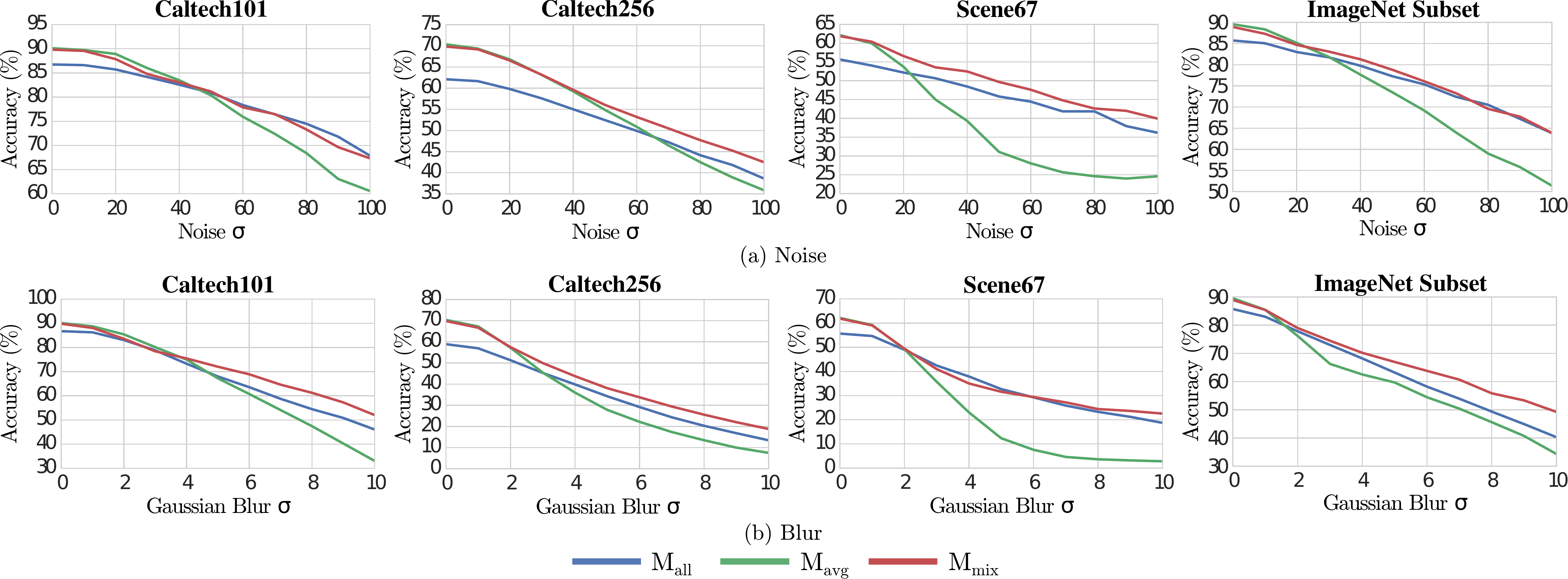}
  \caption{\textbf{Performance of ensemble methods.} The proposed mixture model ($M_{mix}$) achieves good performance over all distortion levels and types compared with an averaging ensemble ($M_{avg}$) and a single model baseline trained on the two distortions ($M_{all}$).}
  \label{fig:ens_result}
\end{figure*}

\begin{table}[!tb]
  \small
  \centering
\begin{tabular}{lccc}
  \toprule
Model & Noise AUC & Blur AUC &  Average \\
  \midrule
  $M_{clean}$ & 0.58 & 0.35 & 0.46  \\
  $M_{noise}$ & 0.79 & 0.40 & 0.59 \\
  $M_{blur}$ & 0.58 & 0.71  & 0.64 \\
  $M_{all}$ & 0.80 & 0.68 & 0.74 \\
  \midrule
  $M_{avg}$ & 0.78 & 0.66 & 0.72  \\
  $M_{max}$ & 0.78 & 0.69 & 0.73  \\
  $M_{hardmix}$ & 0.79 & 0.71 & 0.75 \\
  $M_{mix}$ & \textbf{0.80} & \textbf{0.72} & \textbf{0.76}\\
  \bottomrule
  \end{tabular}
\caption{Caltech101 AUC.}
\label{tab:c101}
\end{table}

\begin{table}[!tb]
  \small
  \centering
\begin{tabular}{lccc}
  \toprule
Model & Noise AUC & Blur AUC & Average \\
  \midrule
  $M_{clean}$ & 0.34 & 0.20 & 0.27  \\
  $M_{noise}$ & 0.54 & 0.19 & 0.36  \\
  $M_{blur}$ & 0.37 & 0.40 & 0.38 \\
  $M_{all}$ & 0.52 & 0.35 & 0.44 \\ 
  \midrule
  $M_{avg}$ & 0.55 & 0.34 & 0.44 \\
  $M_{max}$ & 0.54 & 0.34 & 0.44 \\
  $M_{hardmix}$ & 0.54 & 0.40 & 0.47  \\
  $M_{mix}$ & \textbf{0.57} & \textbf{0.41} & \textbf{0.49} \\
  \bottomrule
  \end{tabular}
\caption{Caltech256 AUC.}
\label{tab:c256}
\end{table}

\paragraph{Results}
We use the same experimental setup as in Section \ref{sec:finetune}. For each distortion type we consider the following models: our proposed mixture model ($M_{mix}$), a hard mixture baseline ($M_{hardmix}$), an averaging ensemble ($M_{avg}$), a max ensemble ($M_{max}$), a single model fine-tuned on both distortion types ($M_{all}$), and the models from Section \ref{sec:finetune} trained on single distortions.

$M_{all}$ is a baseline single model that is fine-tuned on both noisy and blurry images. During training, half of the images in a mini-batch are clean, one quarter are distorted with Gaussian noise, and the remaining quarter are distorted with Gaussian blur.


$M_{avg}$ is an ensemble model that averages the soft-max outputs of the ensemble members. $M_{max}$ gives the class prediction that has the maximum value over all of the soft-max outputs of the models. This will choose the class that is most confidently predicted across the ensemble members.

$M_{hardmix}$ is another baseline ensemble. This baseline is similar to $M_{mix}$, but we assign a weight of 1 to the model corresponding to the largest weight predicted by the gating network. The other ensemble members receive a weight of zero.

Figure \ref{fig:ens_result} shows the performance curves for $M_{mix}$, $M_{avg}$ and $M_{all}$ as a function of distortion levels for the different ensemble models. For most levels and types of distortion, our mixture model achieves the highest accuracy. For lower distortion levels, the mixture gives similar performance as the averaging ensemble, and for higher levels of distortion the mixture achieves similar performance as the model fine-tuned to the particular level of distortion.

We additionally compute the area under the curve (AUC) of the accuracy curves for each distortion type. The AUC is normalized so that 1.00 would achieve 100\% accuracy at all distortion levels. The summary of AUC results is shown in Tables \ref{tab:c101}, \ref{tab:c256}, and \ref{tab:s67}. The average performance of our method is superior to all of the baselines.



\begin{table}[!tb]
  \small
  \centering
\begin{tabular}{lccc}
  \toprule
Model & Noise AUC & Blur AUC & Average \\
  \midrule
  $M_{clean}$ & 0.20 & 0.14 & 0.17 \\
  $M_{noise}$ & 0.49 & 0.17 & 0.33  \\
  $M_{blur}$ & 0.22 & 0.35 & 0.28 \\
  $M_{all}$ & 0.46 & 0.35 & 0.41  \\
  \midrule
  $M_{avg}$ & 0.37 & 0.23 & 0.30  \\
  $M_{max}$ & 0.40 & 0.27 & 0.33  \\
  $M_{hardmix}$ & 0.49 & 0.35 & 0.42  \\ 
  $M_{mix}$ & \textbf{0.50} & \textbf{0.36} & \textbf{0.43} \\
  \bottomrule
  \end{tabular}
\caption{Scene67 AUC.}
\label{tab:s67}
\end{table}

\begin{table}[!tb]
  \small
  \centering
\begin{tabular}{lccc}
  \toprule
Model & Noise AUC & Blur AUC & Average \\
  \midrule
  $M_{clean}$ & 0.52 & 0.28 & 0.40 \\
  $M_{noise}$ & 0.77 & 0.30 & 0.54  \\
  $M_{blur}$ & 0.49 & 0.67 & 0.58 \\
  $M_{all}$ & 0.77 & 0.63 & 0.70  \\
  \midrule
  $M_{avg}$ & 0.72 & 0.60 & 0.66  \\
  $M_{max}$ & 0.73 & 0.63 & 0.67 \\
  $M_{hardmix}$ & 0.77 & 0.67 & 0.72  \\ 
  $M_{mix}$ & \textbf{0.78} & \textbf{0.68} & \textbf{0.73} \\
  \bottomrule
  \end{tabular}
\caption{50 class ImageNet subset AUC.}
\label{tab:imagenet}
\end{table}

\section{Tree Ensembles}
\label{sec:tree}
Although the mixture of experts approach can simultaneously achieve good accuracy on multiple types of distortions, the primary disadvantage is that the ensemble requires approximately 3x the number of parameters as a single model. However, it is likely that some parameters in the ensemble members can be shared. In this section we will explore weight sharing mechanisms that can achieve similar or greater accuracy with much fewer parameters.

\begin{figure*}[!tb]
  \centering
  \includegraphics[width=0.75\textwidth]{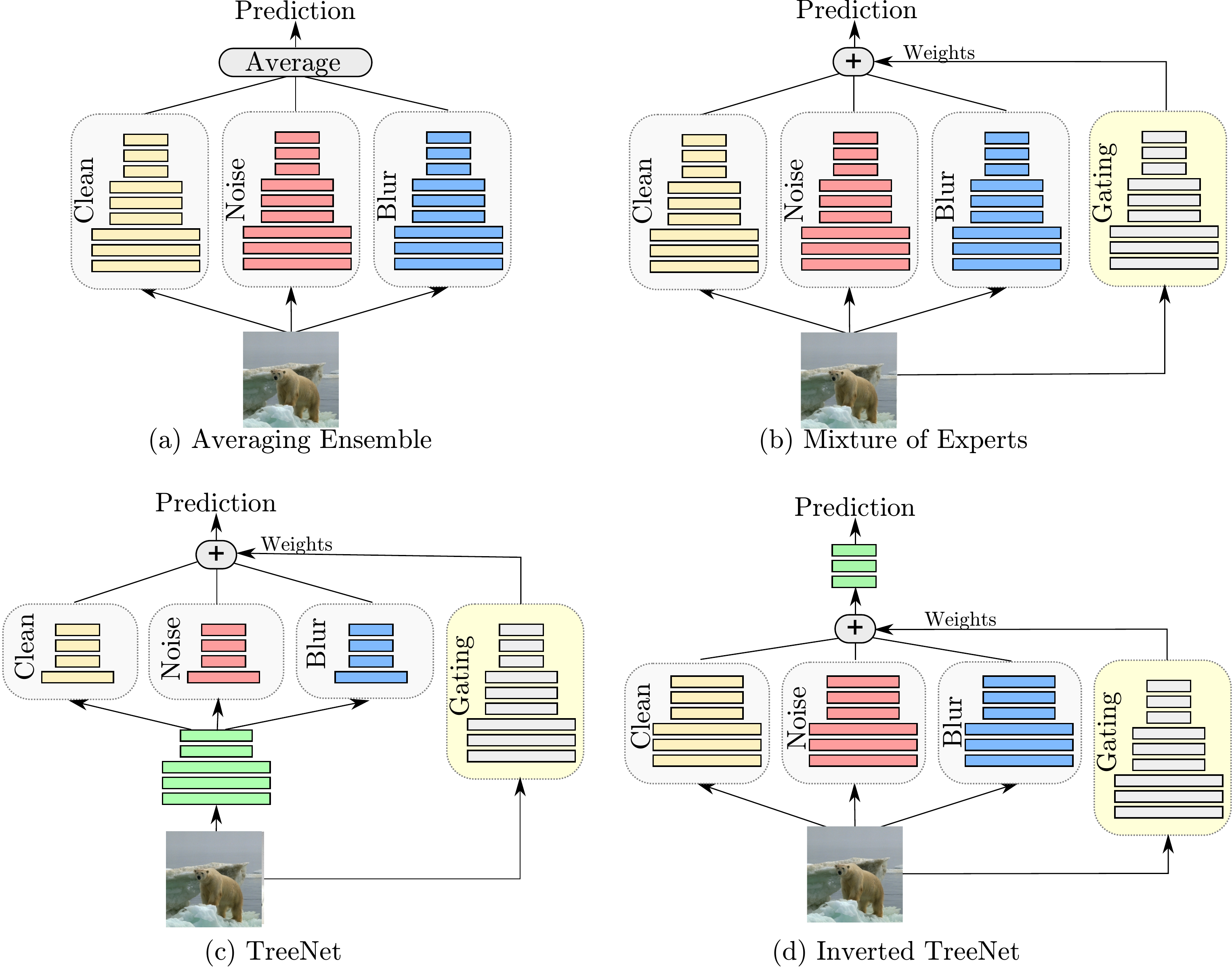}
  \caption{\textbf{Structures of ensemble methods.} Each shaded gray block consists of layers trained on a particular type of distortion. Green layers denote layers that have shared parameters.}
  \label{fig:ensemble}
\end{figure*}

\paragraph{TreeNets}
It is well known that early layers of neural networks often encode similar features \cite{lenc}. Lee \etal propose TreeNets \cite{treenet} that share layer parameters across early layers of members of an ensemble. We consider a similar structure, but instead of simple averaging, we use the gating unit trained in Section \ref{sec:ensemble}. Figure \ref{fig:ensemble}(c) shows a diagram of the TreeNet with gating unit. Each branch of the tree is trained separately using different distortions with the same strategy as in Section \ref{sec:finetune}. For comparison, Figures \ref{fig:ensemble}(a) and \ref{fig:ensemble}(b) show the structures for $M_{avg}$ and $M_{mix}$, which do not incorporate weight sharing.

In training the TreeNet it would be ideal to train the entire model jointly, but jointly training the ensemble has significant memory requirements. Instead we initialize each ensemble member with a model fine-tuned on clean images from the particular dataset. We freeze the parameters of the desired shared layers and fine-tune the other layers for each ensemble member. Since the shared layers do not change during fine-tuning, the layer parameters only need to be stored once and the activations only need to be computed once per-image at test time.

\paragraph{Inverted-TreeNets}
TreeNets, as described in \cite{treenet}, only share the lower layers of a network, because the lower layers are assumed to describe similar features. However, for our particular problem the lower layers may need to adapt differently for the different distortion types. Furthermore, the later fully connected layers have more parameters than the earlier convolutional layers, so sharing later layers would yield more parameter savings.

Thus we propose a network structure called \emph{Inverted-TreeNets} where the parameters of later layers are shared between ensemble members and the early layers can adapt to different distortion types (Figure \ref{fig:ensemble}(d)). We train the Inverted-TreeNets in the same manner as the TreeNet by freezing the shared layers of a network trained on clean images. During testing the weights from the gating network are applied before the first shared layer such that the shared layers are only computed once.

\begin{table}[!tb]
  \small
  \centering
\begin{tabular}{lcccc}
  \toprule
Branch  & \multicolumn{3}{c}{$\overline{AUC}$} & Avg Params \\
Point  & Caltech101 & Caltech256 & Scene67 & (million) \\
  \midrule
  $M_{mix}$ & 0.76 & 0.49 & 0.43 & 404.62 \\
  \midrule
  Conv2\_1 & \textbf{0.77} & \textbf{0.49} & 0.43 & 404.39 \\
  Conv3\_1 & 0.75 & 0.48 & 0.43 & 404.10 \\
  Conv4\_1 & 0.74 & 0.47 & \textbf{0.44} & 401.15 \\
  Conv5\_1 & 0.72 & 0.45 & 0.38 & 389.35 \\
  FC6 & 0.64 & 0.42 & 0.32 & 375.19 \\
  FC7 & 0.60 & 0.40 & 0.31 & 169.66 \\
  FC8 & 0.58 & 0.39 & 0.30 & 136.10 \\
  \bottomrule
  \end{tabular}
\caption{TreeNet branch point vs average AUC and number of unique parameters (averaged across 3 datasets). $\overline{AUC}$ is the average AUC of the curves for blur and noise. The number of parameters includes the gating network (98.5 thousand parameters). Layers below the branch point are shared and layers including and above the branch point are not shared.}
\label{tab:treenet}
\end{table}

\begin{table}[!tb]
\vspace{7pt}
  \small
  \centering
\begin{tabular}{lcccc}
  \toprule
Branch  & \multicolumn{3}{c}{$\overline{AUC}$} & Avg Params \\
Point  & Caltech101 & Caltech256 & Scene67 & (million) \\
  \midrule
  Conv2\_1 & 0.55 & 0.34 & 0.29 & 135.01  \\
  Conv3\_1 & 0.66 & 0.41 & 0.40 & 135.46 \\
  Conv4\_1 & 0.73 & 0.47 & 0.45 & 138.41 \\
  Conv5\_1 & 0.76 & 0.51 & \textbf{0.48} & 150.21 \\
  FC6 & 0.77 & \textbf{0.52} & 0.47 & 164.37 \\
  FC7 & \textbf{0.78} & 0.47 & 0.47 & 369.90 \\
  FC8 & 0.77 & 0.50 & 0.45 & 403.46 \\
  \midrule
  $M_{mix}$ & 0.76 & 0.49 & 0.43 & 404.62 \\
  \bottomrule
  \end{tabular}
\caption{Inverted-TreeNet branch point vs average AUC and number of unique parameters (averaged across 3 datasets). $\overline{AUC}$ is the average AUC of the curves for blur and noise. The number of parameters includes the gating network (98.5 thousand parameters). Layers below the branch point are not shared and layers including and above the branch point are shared. }
\label{tab:inv_treenet}
\end{table}

\paragraph{Results}
The branch point of the TreeNet or Inverted-TreeNet represents a trade-off between model complexity and classification accuracy. In Tables \ref{tab:treenet} and \ref{tab:inv_treenet} we compute the average area under the distortion-accuracy curves for blur and noise ($\overline{AUC}$) and compare it to the total number of parameters of the ensemble for different branch points.

Our results show that the proposed Inverted-TreeNets give a better performance-parameter trade-off than TreeNets. This indicates that the early layers are more important to adapt to the image distortions. The later layers can be identical regardless of the input image distortion. The number of parameters at the FC6 branch point of the Inverted-TreeNet is roughly 40\% of the full mixture model, but surprisingly it achieves greater AUC than the full mixture model. A possible explanation for this improvement is that the networks cannot over-fit the fully connected layers to the distorted data. Additionally, compared with a TreeNet model with a similar number of parameters (e.g., FC7 branch point), the Inverted-TreeNet achieves a much higher average AUC (e.g., 0.77 vs 0.60 on the Caltech101 dataset).

It is not surprising that most of the savings in terms of parameters arise from sharing the fully connected layers, as these layers have the most parameters in the network. Additional parameter savings can be achieved with compression methods such as \cite{deep-compression}, but we note that the tree methods of sharing parameters are orthogonal to compression methods and will still give parameter savings.




\section{Discussion and Conclusion}

We have presented a mixture of experts approach that enables robust image classification across different distortion types and levels. In addition to this, we proposed a new type of weight sharing, Inverted-TreeNets, to significantly reduce the number of parameters of the mixture model, while also achieving greater accuracy. 

Although we have only focused on image classification problems, image distortions could affect any problems that rely on neural networks, such as semantic segmentation \cite{fcn} or object detection \cite{rcnn}. Furthermore, the improvements presented in this work are orthogonal to any improvements due to newer neural network architectures.

Our model predicts appropriate weights based on the characteristics of the input image without explicitly predicting the type or level of the distortion. Further improvements could be made by explicitly incorporating techniques to predict distortion type or level such as \cite{bovik, noise-est}. Finally, although we only test on blur and noise, our model could be easily applied to cases where there are many more distortion types.




{\small
\bibliographystyle{ieee}
\bibliography{refs}
}
\end{document}